# Electricity consumption forecasting method based on MPSO-BP neural network model

Youshan Zhang [1, 2,a], Liangdong Guo[2, b], Qi Li [3, c] and Junhui Li[2, d]

[1] Neuroimaging and Informatics, University of Southern California, Los Angeles, CA, USA

[2] School of Science, University of Science and Technology Liaoning, Anshan, 114051, China

[3] Department of Automation, BOHAI University, Liaoning, Jinzhou, 121013, China

[a] youshanz@usc.edu, [b] ldguo@ustl.edu.cn, [c] liqibaohaiu@163.com, [d] 714978988@163.com

**Abstract:** This paper deals with the problem of the electricity consumption forecasting method. A MPSO-BP(modified particle swarm optimization-back propagation) neural network model is constructed based on the history data of a mineral company of Anshan in China. The simulation showed that the convergence of the algorithm and forecasting accuracy using the obtained model are better than those of other traditional ones, such as BP,PSO, fuzzy neural network and so on. Then we predict the electricity consumption of each month in 2017 based on the MPSO-BP neural network model.

## 1 Introduction

Mineral companies consume large quantities of electricity in the processing of coal every day. The electricity consumption predicting system is always an important part of planning and operating of the power. Because of the complicated change of the electrical power system, it is difficult to establish an exact predicting model [1]. Many companies have changed the traditional methods to predict the electricity consumption, but the accuracy is not high. Traditional BP neural network training algorithms are mostly based on the gradient. The speed of network learning process convergence is slow and falls into the local minimum value easily. It is also difficult to decide the number of neurons in the hidden layer. In terms of the electric power loading randomness, it lacks the ability of precise to screen data processing. The original particle swarm optimization (PSO) has many advantages such as the simple algorithm, easily implement and less parameters. However, it has some disadvantages like is not sensitive to the environmental changes and falls into non-optimal regions easily [2-5].

In this paper, PSO-BP algorithm is modified to train the neural network parameters, realize the optimizing of the network and achieve the automatically optimized parameters of BP neural network. The algorithm is applied to predict the electricity consumption prediction by using Matlab. In addition, our method is used to compare with methods of BP, PSO, Elman, FNN, and ANFIS [6-10], the results show that our algorithm has a higher convergence speed, and it provides a higher accuracy for predicting the electricity consumption.

## 2 Particle Swarm Optimization and Its Improvement

### 2.1 The original particle swarm optimization

In the PSO algorithm, each individual is called a particle, and each particle represents a potential solution. In the D-dimensional search space, each particle is a point in space and group forms by $m$ particles. $\mathbf{z}_i=(z_{i1},z_{i2},\ldots z_{iD})$ and $\mathbf{v}_i=(v_{i1},v_{i2},\ldots v_{id},\ldots,v_{iD})$ are the position vector and the speed vector of $i$ ( $i=1,2,\ldots,m$) particle, $\mathbf{p}_i=(p_{i1},p_{i2},\ldots p_{id},\ldots,p_{iD})$ is the best position of the search particle, $\mathbf{p}_g=(p_{g1},p_{g2},\ldots p_{gd},\ldots,p_{gD})$ is the best position of all particles.

The velocity and position updating equations:

$$v_{id}^{k+1} = v_{id}^k + c_1 r_1^{(k)}(p_{id} - z_{id}^k) + c_2 r_2^{(k)}(p_{gd} - z_{id}^k) \quad (1)$$

$$z_{id}^{k+1} = z_{id}^k + v_{id}^{k+1} \quad (2)$$

Where: $k$ is the iteration index; $r_1$ and $r_2$ are random numbers among [0, 1]; $c_1$ and $c_2$ are the acceleratory coefficient [7].

### 2.2 Modified particle swarm optimization

Modified ideas: 1. Keeping flight diversity of later stage and different flight speed in the same direction. 2. Dividing particles into two categories: high-speed particles satisfy the global search requirements and avoid premature and local optimum; and low-speed particles satisfy the refined search requirements, and avoid exceeding optimal solution. The modified equations are as follows:

$$\begin{cases} v_{id}^{k+1} = \omega v_{id}^k + c_1 r_1 (p_{id} - z_{id}^k) + c_2 r_2 (p_{gd} - z_{id}^k) \\ z_{id}^{k+1} = z_{id}^k + v_{id}^{k+1} \\ v_{id}^n = a(n) v_{id}^0 \quad n = 1, 2, \cdots, j \\ z_{id}^n = z_{id}^{(0)} + z_{id}^n \quad n = 1, 2, \cdots, j \end{cases} \quad (3)$$

where: $v_{id}^{(0)}$ is a base part of the particle $i$ in D-dimensional velocity; $v_{id}^{(m)}$ is a part of the particle $i$ in D-dimensional search speed; $z_{id}^{(0)}$ is a base part of the particle $i$ in D-dimension search position; $z_{id}^{(m)}$ is a part of the particle $i$ in D-dimensional search position; $\omega$ is inertia weight; $P_{id}$ is the best position particle achieved based on its own experience; $P_{gd}$ is the best particle position based on overall swarm's experience.

$$a(n) = \begin{cases} n & v_{id}^{(0)} < N_{i2} \\ n/j & v_{id}^{(0)} \geq N_{i1} \\ 1 \pm n/j & N_{I2} < v_{id}^{(0)} < N_{i1} \end{cases} \quad (4)$$

$a(n)$ is the coefficient variation of speed, $N_{i1}$ is the maximum speed, $N_{i2}$ is the minimum speed, $a(n)$ changes the search speed according to the equation (4).

$$\omega(k) = 2\omega^0 / (1 + \exp(\sigma k / k_{\max})) \quad (5)$$

Where: $\sigma$ is the positive coefficient; $k_{max}$ is the upper limit iteration index; $\omega^0$ is the upper limit of $\omega(k)$; $k$ is the iteration index.

### 2.3 Performance analysis after the improved algorithm

The original PSO algorithm cannot keep the convergence of global optimum, and the probability of getting an optimal solution is small. The modified PSO algorithm introduces a speed variable coefficients $a(m)$ and inertia factor $\omega$. It keeps diversity of particle swarm, and it can obtain the global optimum and improve the convergence speed and accuracy.

## 3 MPSO-BP Neural Network Algorithm

The following equation can be used to judge the fitness of particles:

$$f = \frac{1}{n} \sum_{j=1}^{n} \sum_{k=1}^{m} (d_i - t_k) \quad (6)$$

where $d_i$ is the actual output, $t_k$ is the target output, $m$ is the number of output nodes, and $n$ is the number of training set samples.

### 3.1 Modified particle swarm optimization algorithm realization

1. Initialization: generating the positions and velocities randomly to decide the local best position ($p_{id}$) and the global best position ($p_{gd}$). The equation (3) and (4) decide the initial parameters: $\sigma, \omega, \omega^0, c_1, c_2, r_1, r_2$.

2. Evaluation: calculating the particle fitness function $f$ according to the equation (6), Comparing the best position of each particle with the experience position of each particle, and replacing the current position as the best position if the current position is better than the best position, otherwise, the current position remains unchanged.

3. Update extreme value: comparing the current position of each particle in the group with all the best position experienced. If the position of the particle is better, it will be setting to the best position in the current; otherwise, the position will stay unchanged.

4. Update the inertia weight: The inertia weight is updated according to the equation (5).

5. Update the position and velocity of the particle: the position and velocity of the particle changed by using equation (3) and (4).

6. Check: If the current accuracy not achieves the preset accuracy or the number of iterations not meets the end conditions, turn back to step 2. [11]

Advances in Computer Science Research, volume 50After training the neural network, the electricity consumption can be forecasted.

## 4 MPSO-BP Neural Network Predicting Model

### 4.1 The prediction and simulation of electricity consumption

According to the history data of a mineral company of Anshan in China, the training samples of the neural network are the monthly electricity consumption history data from 2011 to 2014. In addition, the electricity consumption values of 12 months in 2015 are treated as a test sample. By using the training MPSO-BP neural network, the initial parameters are as follows: $\sigma= 0.8$, $\omega=0.7$, $\omega^0=0.9$, $c_1= 2$ and $c_2= 2.4$, the range of $r_1$ and $r_2$ is [0,1], $\alpha= 0.07$, $\eta= 0.80$, $M=50$, the maximum number of iterations is 1000. Different algorithms were used to predict the electricity consumption of a mineral company of Anshan in 2015, the results are shown as following(Due to the length of the space, a part of the results are listed in Table 1)：

Table 1. Energy consumption prediction of each month in 2015

| Methods | | MPSO-BP | | PSO-BP | | BP | | Elman | | FNN | | ANFIS | |
|---|---|---|---|---|---|---|---|---|---|---|---|---|---|
| Months | True value kWh/t | Predict value kWh/t | Relative error % | Predict Value kWh/t | Relative error % | Predict value kWh/t | Relative error % | Predict value kWh/t | Relative error % | Predict Value kWh/t | Relative error % | Predict Value kWh/t | Relative error % |
| 1 | 36.82 | 36.18 | 1.739 | 36.08 | 2.000 | 34.90 | -1.923 | 36.20 | 1.673 | 37.13 | -0.845 | 36.16 | 1.893 |
| 2 | 36.87 | 35.86 | 2.734 | 36.47 | 1.091 | 36.86 | -0.013 | 35.82 | 2.8455 | 37.95 | -2.941 | 36.38 | -3.359 |
| 3 | 36.84 | 36.49 | 0.947 | 36.22 | 1.691 | 35.96 | -0.880 | 35.63 | 3.2828 | 36.24 | 1.632 | 36.37 | -3.284 |
| 4 | 35.16 | 36.19 | -2.930 | 36.10 | -2.673 | 34.15 | -1.008 | 35.03 | 0.378 | 35.80 | -1.833 | 34.79 | 1.707 |

Table 2. Electric energy error of each month in 2015

| Methods | MPSO-BP | PSO-BP | BP | Elman | FNN | ANFIS |
|---|---|---|---|---|---|---|
| MSE | 0.966627 | 1.045105 | 4.273972 | 2.297225 | 0.9128 | 0.809817 |
| Average relative error% | 0.9993 | 2.2728 | 4.7355 | 2.9997 | 2.0883 | 2.883 |

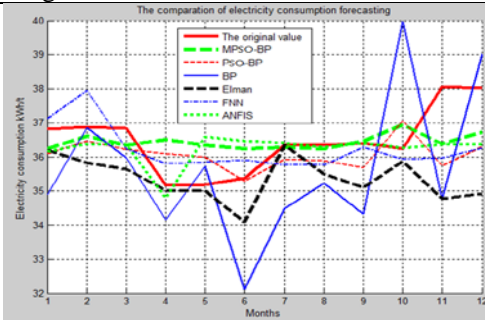
Fig. 1 Electricity consumption prediction

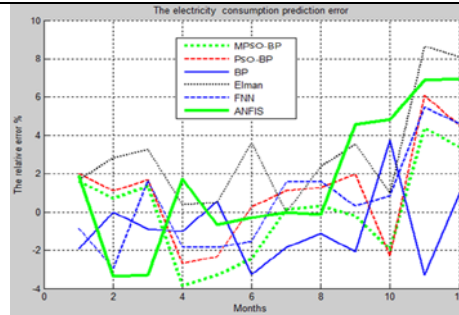
Fig.2 The prediction error

It can be seen from table 1, table 2, figure 1, and figure 2 that the predicting value of the electricity consumption of MPSO-BP algorithm is very close to the actual value. Therefore, this method has a higher accuracy than the others. Figure 3 can also show that the predicting results of Elman and BP algorithm are easy to appear oscillating. However, MPSO-BP predicting model has a fast training speed, high precision, and a good effect. The changing of relative error is also uniform, and the changing range is between 0% with 4.5%．

### 4.2 The analysis of simulation results

In order to evaluate the effect of MPSO-BP prediction model, PSO-BP, BP, Elman, FNN and ANFIS predicting models are trained and compared with the results of MPSO-BP model. The results are shown in Table 3:

Table 3 Comparison of model training performance

| Model | Accuracy | Iterations | Average relative error | Maximum relative error |
|---|---|---|---|---|
| ANFIS | 0.005 | 4800 | 2.8834% | 3.7342% |
| FNN | 0.005 | 4560 | 2.4927% | 3.1275% |
| Elman | 0.005 | 1500 | 3.6764% | 4.4959% |
| BP | 0.005 | 2000 | 4.2437% | 7.0630% |
| PSO-BP | 0.005 | 500 | 1.8223% | 2.3154% |
| MPSO-BP | 0.005 | 200 | 0.8677% | 2.0535% |

As can be seen from table 3, the error and the number of iterations of MPSO-BP neural network are obviously less than the traditional ANFIS, FNN, Elman, BP, and PSO-BP neural networks. MPSO-BP neural network has faster convergence, and higher convergence precision, so adopting this optimization of the calculation is feasible.

Adopting the trained MPSO-BP neural network forecasting model, selected five months randomly to predict the electricity consumption, and the results shown in Table 4:

Table 4. Electricity consumption forecasting accuracy

| time | 8/2011 | 11/2012 | 12/2013 | 4/2014 | 7/2015 |
|---|---|---|---|---|---|
| Original value(kWh/t) | 36.22 | 36.67 | 34.23 | 36.36 | 35.38 |
| Predictive value(kWh/t) | 36.10 | 36.76 | 34.03 | 36.86 | 35.34 |
| Accuracy (%) | 99.7 | 99.8 | 99.4 | 98.6 | 99.9 |

It has shown from the table 4, the predicting accuracy can reach about 99%, and the highest predicting accuracy was 99.9%. Obviously, for multiple factors in the predicting of electricity consumption, such accuracy is difficult to achieve. It shows that the MPSO-BP predicting model can achieve high predicting accuracy and high practicality.

**4.3 The electricity consumption results**

Based on the MPSO-BP neural network model, we predict the monthly electricity consumption of the mineral company in 2017, the results are: 34.52, 36.32, 35.44, 36.01, 34.96, 35.62, 37.03, 35.15, 37.41, 35.92, 37.16, and 37.96 kWh/t.

**5 Conclusions**

In this paper, we proposed a modified PSO-BP neural network method to predict the electricity consumption. The MPSO-BP neural network algorithm is used to train the network, and achieve the optimization of BP network parameter. This algorithm analysis shows that the MPSO-BP algorithm overcomes problems of BP neural network and traditional particle swarm optimization algorithm. This improved algorithm has a good predictive ability, improves the predicting accuracy and provides a feasible and practical method for forecasting electricity consumption.